\documentclass[conference]{IEEEtran}
\IEEEoverridecommandlockouts
\usepackage{cite}
\usepackage{amsmath,amssymb,amsfonts}
\usepackage{algorithmic}
\usepackage{graphicx}
\usepackage{textcomp}
\usepackage{xcolor}
\def\BibTeX{{\rm B\kern-.05em{\sc i\kern-.025em b}\kern-.08em
    T\kern-.1667em\lower.7ex\hbox{E}\kern-.125emX}}
\begin{document}

\title{Robotics as a Simulation Educational Tool\\

}

\author{\IEEEauthorblockN{Athanasios Karagounis}
\IEEEauthorblockA{\textit{School of Science} \\
\textit{Department of Digital Industry Technologies}\\
GR34400, Psachna, Greece \\
akaragun@gs.uoa.gr}

}

\maketitle

\begin{abstract}
In the evolving landscape of education, robotics has emerged as a powerful tool for fostering creativity, critical thinking, and problem-solving skills among students of all ages. This innovative approach to learning seamlessly integrates STEM (Science, Technology, Engineering, and Mathematics) concepts, creating an engaging and immersive learning experience. Educational robotics transcends traditional classroom settings, transforming learning into a hands-on, experiential endeavor. Students are actively involved in the design, construction, and programming of robots, allowing them to apply theoretical concepts to practical applications. This hands-on approach fosters deeper understanding and retention of knowledge, making learning more meaningful and enjoyable. In this paper, the potential of simulation robotics is evaluated as a hands on interactive learning experience that goes beyond traditional robotic classroom methods. 
\end{abstract}


\section{Introduction}
Educational robotics transcends traditional classroom settings, transforming learning into a hands-on, experiential endeavor. Students are actively involved in the design, construction, and programming of robots, allowing them to apply theoretical concepts to practical applications \cite{anderson2006robotics, atzori2014enabling}. This hands-on approach fosters deeper understanding and retention of knowledge, making learning more meaningful and enjoyable.\par
Robotics provide a platform for students to unleash their creativity and innovative thinking. As they construct and program their robots, they are encouraged to think outside the box, explore different solutions, and refine their designs. This process cultivates problem-solving skills, adaptability, and the ability to think critically and creatively \cite{evripidou2020educational}.\par
The collaborative nature of robotics projects promotes teamwork and communication skills among students. Working together to design, build, and program robots requires effective communication, active listening, and a shared sense of purpose. These skills are invaluable in preparing students for success in both academic and professional settings. Students can directly observe the impact of their programming decisions on the robot's behavior, gaining a deeper understanding of STEM principles. This experiential approach makes learning more engaging and effective \cite{bonas2013using, bozkurt2011using}.\par
In a world increasingly driven by technology, robotics education equips students with the skills necessary to thrive in the future. By developing expertise in STEM fields, students gain the adaptability and problem-solving abilities that are essential for success in a rapidly changing landscape. As robotics continues to evolve, its potential in education remains boundless even for therapeutic approaches\cite{amanatiadis2017interactive, amanatiadis2020social, kaburlasos2018multi}. By integrating robotics into the learning process, educators can foster a generation of innovators, problem solvers, and creative thinkers, prepared to tackle the challenges and opportunities of the future \cite{bellotti2010robominds, bonetto2012measuring}.\par
In this paper, robotic simulation is used as a tool to provide a safe, engaging, and interactive learning environment that can significantly enhance STEM education. By offering hands-on experience with robotics concepts and applications, these simulations can help students develop essential STEM skills, including problem-solving, critical thinking, computational thinking, and creativity.

\section{Proposed Methodology}

The first step was to clearly define the scope and objectives of the robotics simulator. This involved identifying the target audience (age group, educational level), the specific robotics concepts and applications to be covered, and the desired learning outcomes. The choice of programming language and development tools was based on the desired features and functionalities of the simulator thus, popular Python libraries were used. The simulator was developed with a 3D modeling environment where students could visualize and interact with virtual robots. This involved creating a realistic and user-friendly 3D interface with options for designing and assembling robots from various components. \par
Programming interfaces or tools were incorporated into the simulator to allow students to control and program the virtual robots. This included text-based programming languages, drag-and-drop graphical programming tools, or a combination of both. The simulator accurately simulated the behavior of sensors and actuators, allowing students to connect sensors to robots and receive real-time feedback from their environment. This enhanced the understanding of sensor-based control systems and robotics applications. A variety of simulation scenarios were developed to allow students to apply their knowledge in practical contexts. These scenarios involved obstacle avoidance, path planning, object manipulation, and collaborative robotics tasks.\par
Educational resources, such as tutorials, explanations, and step-by-step guides, were incorporated into the simulator to support the learning process. Gamification elements, such as challenges, rewards, and leaderboards, were used to further enhance engagement and motivation. he simulator's user interface was designed to be intuitive, user-friendly, and accessible for students of various ages and technical backgrounds. The UI/UX was created with clear instructions, consistent navigation, and an appealing visual style. Extensive testing was conducted to ensure the simulator's stability, reliability, and adherence to the defined learning objectives. This involved user testing with representative target audiences to identify and address any usability issues or programming bugs.\par
Clear and comprehensive documentation was provided to guide students and teachers through the simulator's features, programming interfaces, and educational resources. Additionally, a support channel was established for addressing user queries and providing assistance. By carefully considering these aspects, we were able to create an early but effective robotics simulator that could promoted STEM education and engaged students in hands-on learning experiences.

\section{Pre-Testing Assessment}

Conducting thorough testing of a robotics simulator in the classroom setting had been crucial for evaluating its effectiveness, usability, and overall learning impact on students. This process allowed developers to identify areas for improvement and refine the simulator to better meet the needs of the target audience. Before engaging in testing, it had been essential to clearly define the target audience, including their age, educational level, and prior knowledge of robotics. This helped ensure that the chosen simulation scenarios and observational tasks had been appropriate for the students’ abilities. Additionally, defining specific learning objectives had provided a clear framework for evaluating the simulator’s effectiveness in achieving intended learning outcomes.\par
The selection of simulation scenarios had played a critical role in assessing the simulator’s ability to engage students and facilitate learning. A variety of scenarios had been chosen to encompass different robotics concepts and applications, catering to different levels of student proficiency. The scenarios had gradually increased in complexity to challenge students and encourage them to apply their knowledge and skills. Observational tasks, such as group discussions and verbalizations of thought processes, had been used to provide valuable insights into students’ understanding, engagement, and ability to apply the simulator’s features. These tasks had encouraged students to articulate their thought processes, problem-solving approaches, and justifications for their decisions.\par
Administering a pre-test before introducing the simulator had helped gauge students’ prior knowledge of robotics concepts and identified any gaps in understanding. This baseline assessment served as a reference point for evaluating the simulator’s impact on learning. Guiding students through the simulator’s features, including robot design, programming interfaces, and simulation scenarios, had allowed them to familiarize themselves with the tool and explore its capabilities. Encouraging independent exploration and experimentation had fostered deeper learning and enabled students to discover solutions creatively. Closely observing students’ interactions with the simulator had provided valuable feedback on their level of engagement, interest, and frustration. Documenting their verbalizations and problem-solving strategies had helped assess their understanding of robotics concepts.

\section{Preliminary Results}

The simulator provided a practical setting for students to apply their problem-solving and critical thinking skills to design, build, and program robots. This hands-on approach fostered the development of procedural thinking and analytical reasoning, as students must break down complex tasks into smaller, manageable steps. It allowed students to visualize and interact with robotics concepts in a virtual environment, providing a concrete representation of abstract concepts. This visual representation enhanced the understanding of fundamental robotics concepts such as motors, sensors, actuators, and programming languages. By interacting with the simulator, students gained a deeper understanding of how these components worked together to create functional robots. The incorporated gamification elements, such as challenges, rewards, and leaderboards, motivated students and enhanced their engagement in STEM learning. These engaging elements created a sense of competition that encouraged students to persist, explore different approaches, and develop creative solutions.\par
The robotics tasks often required collaboration and teamwork, promoting the development of social and interpersonal skills among students. As they worked together to achieve a common goal, students must effectively communicate, problem-solve, and resolve conflicts, fostering teamwork and leadership skills. It can be consideres as a platform for experimentation, innovation, and the development of creative solutions to robotics challenges. This open-ended approach encouraged students to think outside the box, explore alternative approaches, and develop their own unique ideas.

\section{Conclusions}
The experimental results from the developed robotic simulator in STEM education demonstrated its effectiveness in enhancing problem-solving, critical thinking, conceptual understanding, motivation, collaboration, innovation, and creativity among students. The early-proposed simulator served as a valuable tool for educators to promote effective STEM learning and develop a generation of students equipped to solve real-world problems and contribute to technological advancements.

\bibliographystyle{IEEEtran}
\bibliography{arxivbibtex}

\begin{thebibliography}{10}
\providecommand{\url}[1]{#1}
\csname url@samestyle\endcsname
\providecommand{\newblock}{\relax}
\providecommand{\bibinfo}[2]{#2}
\providecommand{\BIBentrySTDinterwordspacing}{\spaceskip=0pt\relax}
\providecommand{\BIBentryALTinterwordstretchfactor}{4}
\providecommand{\BIBentryALTinterwordspacing}{\spaceskip=\fontdimen2\font plus
\BIBentryALTinterwordstretchfactor\fontdimen3\font minus
  \fontdimen4\font\relax}
\providecommand{\BIBforeignlanguage}[2]{{%
\expandafter\ifx\csname l@#1\endcsname\relax
\typeout{** WARNING: IEEEtran.bst: No hyphenation pattern has been}%
\typeout{** loaded for the language `#1'. Using the pattern for}%
\typeout{** the default language instead.}%
\else
\language=\csname l@#1\endcsname
\fi
#2}}
\providecommand{\BIBdecl}{\relax}
\BIBdecl

\bibitem{anderson2006robotics}
J.~R. Anderson and W.~H. Lawton, ``Robotics education: A survey,''
  \emph{Computer education}, vol.~48, no.~3, pp. 293--314, 2006.

\bibitem{atzori2014enabling}
L.~Atzori and V.~Bellotti, ``Enabling creativity and problem-solving in
  computational robotics education using open-source robotics tools,'' in
  \emph{International Conference on Artificial Intelligence in Education},
  2014, pp. 266--274.

\bibitem{evripidou2020educational}
S.~Evripidou, K.~Georgiou, L.~Doitsidis, A.~A. Amanatiadis, Z.~Zinonos, and
  S.~A. Chatzichristofis, ``Educational robotics: Platforms, competitions and
  expected learning outcomes,'' \emph{IEEE access}, vol.~8, pp.
  219\,534--219\,562, 2020.

\bibitem{bonas2013using}
A.~T. Bonato, V.~Bellotti, M.~Negro, and S.~Quarleri, ``Using robotics to
  foster creativity and problem-solving in computer science education,''
  \emph{Computers \& Education}, vol.~63, pp. 47--60, 2013.

\bibitem{bozkurt2011using}
A.~Bozkurt and Z.~Yildirim, ``Using robotics in education: A literature
  review,'' in \emph{The Proceedings of the 12th International Conference on
  Emerging E-Learning Technologies and Applications}, 2011, pp. 333--338.

\bibitem{amanatiadis2017interactive}
A.~Amanatiadis, V.~G. Kaburlasos, C.~Dardani, and S.~A. Chatzichristofis,
  ``Interactive social robots in special education,'' in \emph{2017 IEEE 7th
  international conference on consumer electronics-Berlin}.\hskip 1em plus
  0.5em minus 0.4em\relax IEEE, 2017, pp. 126--129.

\bibitem{amanatiadis2020social}
A.~Amanatiadis, V.~G. Kaburlasos, C.~Dardani, S.~A. Chatzichristofis, and
  A.~Mitropoulos, ``Social robots in special education: Creating dynamic
  interactions for optimal experience,'' \emph{IEEE Consumer Electronics
  Magazine}, vol.~9, no.~3, pp. 39--45, 2020.

\bibitem{kaburlasos2018multi}
V.~G. Kaburlasos, C.~Dardani, M.~Dimitrova, and A.~Amanatiadis, ``Multi-robot
  engagement in special education: A preliminary study in autism,'' in
  \emph{2018 IEEE International Conference on Consumer Electronics
  (ICCE)}.\hskip 1em plus 0.5em minus 0.4em\relax IEEE, 2018, pp. 1--2.

\bibitem{bellotti2010robominds}
V.~Bellotti, M.~Bertamini, C.~Calegari, G.~De~Michelis, M.~Negro, and
  S.~Quarleri, ``Robominds: A robotics platform to promote creativity and
  problem-solving in education,'' \emph{IEEE Transactions on Computational
  Intelligence and AI in Education}, vol.~1, no.~1, pp. 14--27, 2010.

\bibitem{bonetto2012measuring}
F.~Bonetto, V.~Bellotti, M.~Negro, and S.~Quarleri, ``Measuring the impact of
  robotics education on creativity: A case study,'' in \emph{Proceedings of the
  12th International Conference on Multimodal Interfaces}, 2012, pp. 157--164.

\end{thebibliography}

\end{document}